\documentclass[times,twocolumn,final,authoryear]{elsarticle}

\usepackage{prletters}
\usepackage{framed,multirow}

\usepackage{amssymb}
\usepackage{latexsym}

\usepackage{url}
\usepackage{xcolor}
\definecolor{newcolor}{rgb}{.8,.349,.1}

\journal{Pattern Recognition Letters}

\begin{document}

\begin{frontmatter}

\title{Image-to-Video Person Re-Identification by Reusing Cross-modal Embeddings}

\author[1]{Zhongwei \snm{Xie}} 
\cortext[cor1]{Corresponding author: Lin Li
  Tel.: +86-27-87216780;  
  fax: +86-27-87216780;}
\ead{cathylilin@whut.edu.cn}
\author[1]{Lin \snm{Li}\corref{cor1}}
\author[1]{Xian \snm{Zhong}}
\author[1]{Luo \snm{Zhong}}

\address[1]{School of Computer Science and Technology, Wuhan University of Technology, Wuhan and 430070, China}

\received{1 May 2013}
\finalform{10 May 2013}
\accepted{13 May 2013}
\availableonline{15 May 2013}
\communicated{S. Sarkar}

\begin{abstract}
Image-to-video person re-identification identifies a target person by a probe image from quantities of pedestrian videos captured by non-overlapping cameras. Despite the great progress achieved, it's still challenging to match in the multimodal scenario, i.e. between image and video. Currently, state-of-the-art approaches mainly focus on the task-specific data, neglecting the extra information on the different but related tasks. In this paper, we propose an end-to-end neural network framework for image-to-video person re-identification by leveraging cross-modal embeddings learned from extra information. Concretely speaking, cross-modal embeddings from image captioning and video captioning models are reused to help learned features be projected into a coordinated space, where similarity can be directly computed. Besides, training steps from fixed model reuse approach are integrated into our framework, which can incorporate beneficial information and eventually make the target networks independent of existing models. Apart from that, our proposed framework resorts to CNNs and LSTMs for extracting visual and spatiotemporal features, and combines the strengths of identification and verification model to improve the discriminative ability of the learned feature. The experimental results demonstrate the effectiveness of our framework on narrowing down the gap between heterogeneous data and obtaining observable improvement in image-to-video person re-identification.

\end{abstract}


\end{frontmatter}


\section{Introduction}
\label{intro}
With widespread use of surveillance cameras and enhanced awareness of public security, person re-identification(re-id), the task of recognizing people in a non-overlapping camera network, has attracted especial attention of computer vision and pattern recognition communities\citep{23,24}. Given an image/video of a person-of-interest(query), person identification identifies the person from images/videos taken from a different camera. After many years of great efforts, person re-id still remains a notably challenging task. The main reason is that a person's appearance will dramatically change across camera views due to the large variations in illumination, poses, viewpoints and cluttered background\citep{1}. 

According to the scenarios of the re-identification, existing person re-identification works can be roughly divided into two categories: image-based and video-based person re-identification. The former focuses on the matching between a probe image of one person and the image of the person with the same ID in the gallery sets, which is mainly based on image content analysis and matching, while the latter focuses on the matching between video and video, which can exploit different information such as temporal and motion. A gallery is a collection of images cropped from a different view of unknown people. In both kinds of approaches, the two objects to be matched are homogeneous. 

However, in many practical cases, benefiting from the more and more surveillance cameras installed in public places, person re-identification needs to be conducted between image and video. For example, given an image of criminal suspect, the police want to rapidly locate and track suspect from masses of city surveillance videos. The re-identification under this scenario is called image-to-video person re-identification, where the probe is an image and the gallery consists of videos captured by non-overlapping cameras.

Although more information can be obtained from videos, image-to-video re-identification shares the same common challenges with image-based and video-based person re-identification(e.g. similar appearance, low resolution, large variation in poses, occlusion and different viewpoints) and exists its own difficulty: how to match between two different modalities, i.e. image and video. Image and video are usually represented with different features. Specifically speaking, both visual features and spatiotemporal features can be extracted from a video while only visual features can be obtained from a single image.

Recently, convolutional neural networks(CNN) has shown potential for learning state-of-the-art image feature embedding\citep{2, 3} and recurrent neural networks(RNN) yield promising performance in obtaining spatiotemporal feature from video\citep{4,5}. In general, there are two major types of deep learning structures for person re-identification, i.e. verification models and identification models. Verification models take a pair of data as input and determine whether they belong to the same person or not, which only leverage weak re-id labels and can be regarded as a binary-class classification or similarity regression task\citep{6}, while identification models aim at feature learning by treating person re-identification as a multi-class classification task\citep{3}, but lack direct similarity measurement between input pairs. Due to the complementary advantages and limitations, the two models are combined to improve the performance\citep{10}. 

However, in the image-to-video person re-identification task, a cross-modal system, directly using deep neural networks and the information provided by target task still cannot perfectly bridge the ``media gap'', which means that representations of different modalities are inconsistent. There have been efforts~\citep{7,8} reported to reuse existing models trained mostly for other tasks to construct a new model, which inspires us that presumably we can reuse the extra information learned from the different but related task to help cross the ``media gap''. For most of the works, they directly use the weights from pre-trained deep networks as initial values for the target model, and kick off the pre-trained network structure that in fact can help train the new deep model. Currently, Fixed Model Reuse(FMR) is proposed in~\cite{9} to incorporate the helpful fixed model into the training of a new convolutional model.

In order to address the aforementioned limitations, we propose a novel end-to-end neural network framework for image-to-video person re-identification by reusing cross-model embeddings from different but related tasks. Concretely speaking, the proposed framework consists of convolutional neural networks and long short term memory networks(LSTM) for image feature extraction and video spatiotemporal feature extraction, and combines the strengths of identification model and verification model to improve the discriminative ability of the learned feature representation. Furthermore, cross-modal embeddings, i.e. image-to-text and video-to-text embedding layers are reused to help image-to-video person re-id narrow down the ``media gap'' and improve the performance. 

Our main contributions can be summarized as follows. 

\begin{itemize}

\item [$\bullet$]We propose an end-to-end cross-modal framework for image-to-video re-id, which leverages CNN and LSTM to extract the visual features and spatiotemporal motion features. Additionally, identification loss and verification loss are combined so that the framework can simultaneously learn discriminative feature representations and a similarity metric.  

\item [$\bullet$]Cross-modal embedding from different but related tasks are reused in our proposed framework, i.e. image-to-text and video-to-text embeddings. Experimental results demonstrate that making text be the intermediate between image and video before the similarity learning may help narrow down the ``media gap" and improve pedestrian retrieval accuracy. 

\end{itemize} 

The remainder of this paper is organized as follows. Section~\ref{related_work} briefly reviews some related work about person re-identification in recent years. Then we elaborate on the proposed framework for image-to-video person re-id and present its each part at length in Section~\ref{framework}, followed by experiments in Section~\ref{exp} together with conclusions and future work in Section~\ref{conclusion}. 

\section{Related work}
\label{related_work}

In this section, we briefly introduce some previous works relevant to person re-identification and the framework proposed in this paper.

\subsection{Feature representation for person re-identification}

According to the way of feature representation for person re-identification, the existing literature can be roughly classified into two categories: hand-crafted feature and deep learning based feature.

Commonly, the most popular hand-crafted feature representation for person re-identification are color and texture, such as local binary patterns(LBP), color histograms, local maximal occurrence(LOMO). Color histograms are extracted from HSV and YUV spaces in \citep{11}. Li et al. extract local color descriptor from patches and use hierarchical Gaussianization to capture spatial features\citep{12}. Liao et al. propose the LOMO descriptor based on color and SILTP histograms\citep{13}. \citep{35} leverages an optimized local graph structure to obtain the view-specific feature representation. \citep{37} adopts factorized latent data-cluster representations during each iteration of each view. And \citep{14} extracts the HSV histogram, gradient histogram and the LBP histogram for each local patch.

In recent years, deep learning methods have produced overwhelmingly superior performance in computer vision due to its power in learning discriminative features from large-scale image data\citep{15, 25}. Schroff et al. adopted a deep CNN to learn the image Euclidean embedding. \citep{16} incorporates a multi-level  encoding layer into the CNN networks and obtained video descriptors of varying sequence lengths. \citep{18} proposes a siamese attention architecture that jointly learns spatiotemporal video representations. \citep{17} presents a deep attention-based spatially recursive model to encode critical object parts into spatially expressive representations. Color and optical flow information are leveraged to capture appearance and motion representation\citep{4}. \citep{5} builds RNN-based sequence level representation from a temporally ordered sequence of frame-wise feature.

In our image-to-video person re-id framework, visual features are extracted from each frame of video and probe images by CNNs at first and then the frame features from all time-steps are combined using LSTM networks to give an overall appearance and motion feature for the complete sequence. 

\subsection{Verification-identification models for person re-identification}

Verification models usually take a pair of input data and calculate a similarity score between low-dimensional features to determine whether they belong to the same person, which can be penalized by the contrastive loss. \cite{6} extract pedestrian image features by three part-CNNs and compute cosine distance of the images features as their similarity. \citep{19} adds a patch-matching layer to find similar locations and penalizes the similarity regression by softmax loss. What's more, \citep{20} takes triplet samples as input so that images from the same person and the different person can be considered in the mean time. However, verification models only take advantage of weak re-id labels and lack the consideration of all the annotated information.

In the attempt to take all the re-id labels into consideration, treating person re-id as a multi-class recognition task is employed for feature learning. Thanks to the large-scale training sets provided by recent datasets, training a deep classification model for identification task without overfitting can be possible. \citep{21} combines CNN embeddings with the hand-crafted features in the fully connected layer. \citep{3} builds a classification mtodel based on multiple datasets and presents a new dropout function to identify hundreds of classes. \citep{22} proposes a fully automated person re-identification, which employs KDES as a descriptor and SVM as a classifier. The major problem in the identification models is that it doesn't account for the similarity measurement during the testing pedestrian retrieval procedure.

More recently, some works train the network with the verification loss and identification loss, but they mainly focus on homogenous data. \citep{10} sheds light on combine the verification model and identification model to learn more discriminative pedestrian descriptors for image-based person re-id. \citep{4}makes use of color and optical flow information to capture appearance and motion information for video-based re-id and adopts siamese cost and identification cost to train the model. In this paper, we leverage the complementary nature of two models to improve the discriminative ability of the learned representations for the image-to-video person re-id task.

\subsection{Model reuse approaches}

Model reuse usually attempts to build a model by means of reusing existing available models for other tasks, rather than from scratch\citep{9}. It offers a great potential to reduce the required amount of training examples and training time cost, owning to the good basis for the training of new model set by the exploitation of existing models. \citep{8} initializes a network with weights from pre-trained networks. \citep{7} proposes a new network architecture to transfer features from pre-trained networks. These models generally re-trains on dataset B with pre-trained deep networks on dataset A, i.e., they mainly focus on the transfer of the latent weights and ignores the network structure of pre-trained networks. Currently, Fixed Model Reuse approach(FMR), a more thorough model reuse scheme has been proposed, which arranges the convolution layers and the model or features for general tasks in parallel, using weight propagation(WP), knockdown(KD) and weight propagation after knockdown(WP/K) to train the combined networks and gradually reduces the dependencies between the fixed model/features and the original network.

In our proposed framework, based on the assumption that text can be a intermediate that connecting image and video modality and narrow down the ``media gap'', we adopt the FMR training steps(WP, KD and WP/K) to reuse the cross-modal embedding layers in the image captioning model and video captioning model, which facilitates the similarity learning and improves pedestrian retrieval accuracy.

\section{Proposed framework}
\label{framework}

When dealing with the image-to-video person re-identification task, two major challenges will be inevitably encountered , i.e. how to represent the features of image and video, and how to train the cross-modal similarity learning, which are crucial to the re-id accuracy. In order to address this problem, we proposed an end-to-end framework by reusing by reusing cross-model embeddings from different but related tasks.

In this section, our proposed framework for image-to-video person re-identification is elaborated on mainly in two parts, which are feature representation and verification-identification sub-network, as shown in Fig. 1. The first part is used for extracting the feature representations of image and video, while the second is for improving the discriminative ability of learned features and training a similarity metric between image and video modality, which can optimize the feature representation and similarity learning simutaneously.

\begin{figure*}[!t]
\label{F1}
\centering
\includegraphics[scale=.5]{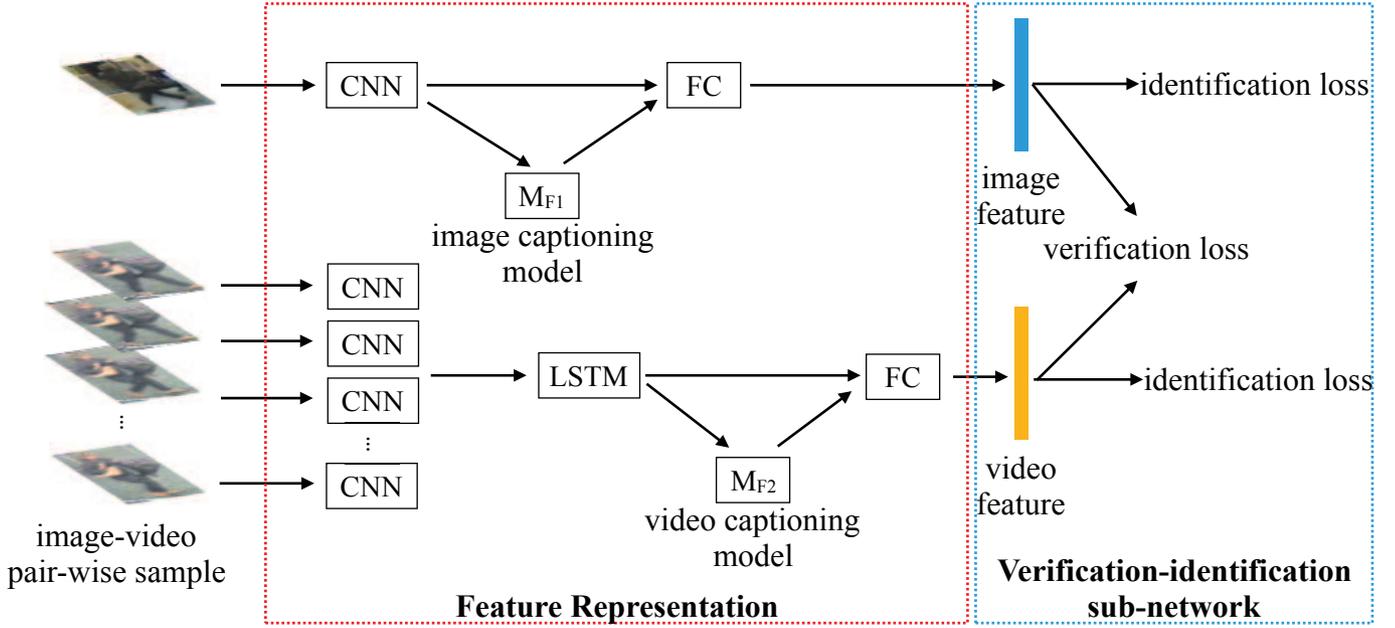}
\caption{The pipeline of proposed framework for image-to-video person re-identification}
\end{figure*}

\subsection{Feature representation}

Several studies have proved that using CNNs is effective in image-based person re-identification task\citep{15,19}. Nonetheless, these CNN models only can deal with single-shot image, not applicable to the arbitrary-length video tracklets in the image-to-video re-id. Fortunately, RNNs, particularly LSTMs, have shown promising stable and powerful performance in modeling long-range dependencies in sequence learning tasks.  Therefore, we decide to combine CNNs and LSTM networks for learning the feature representations of image and video in this work(marked with the red dashed box in Fig. 1).

When extracting the feature representation of input image, it is commonly believed that the superior performance of deep networks is owning to hierarchical feature extraction. Thus,  we adopt CNNs to obtain the image feature for understanding image content, by means of a hierarchy of trainable convolution and pooling operations. Specifically speaking, pre-trained Inception v3 networks\citep{26} without the fully-connected layers are leveraged to generate the image representation due to its fewer parameters and remarkable ability to capture visual features. 

As for the input video, at first, arbitrary-length frames of the video sequence are passed through the CNNs in the similar fashion separately to extract the visual information as a high-level feature representation. Then, each feature of frame is passed forward to the LSTM layer to explore the spatiotemporal information in the video, where it is projected into a low-dimensional feature space and combined with information from previous time steps. Due to the time-series of frames contained in the input video, the use of LSTM layers can help to better capture temporal information presented in the video by passing frame information between time-steps. The LSTMs can easily memorize the long-term dependencies in sequential data, and then the temporal dependency learning can be conveniently converted into the spatial domain. In this work, we follow the similar architecture of \citep{27} for video feature learning.

Resorting to CNNs and LSTMs, the feature representations of input image and video can be obtained. However, they are heterogeneous and inappropriate to directly be dealt with together or matched. For purpose of tackling this difficulty, text modality is brought in to act as an intermediate connecting image and video by reusing the cross-modal embedding layers in the existing image captioning model\footnote{https://github.com/yashk2810/Image-Captioning}\citep{28} and video captioning model\footnote{https://github.com/vsubhashini/caffe/tree/recurrent/examples/s2vt}\citep{29}, which are marked with $M_{F1}$ and $M_{F2}$ respectively in Fig. 1. The training steps(weight propagation, knockdown and weight propagation after knockdown) in the fixed model reuse approach\citep{9} are adopted to reduce the dependencies between the ``fixed'' structure and our target network, and make the network consistent.

\subsection{Verification-identification sub-network}
On the one hand, verification model merely makes use of the weak labels so as to take limited relationships into consideration. On the other hand, identification model does not account for the similarity measurement between input pairs. Therefore, we adopt the verification-identification model(shown in the blue dashed box in Fig. 1) to leverage their complementary nature and combine their respective strengths to improve the discriminative ability of learned features and train a similarity metric between image and video modality.

Similar to the approach \citep{30} for face recognition, we train the verification-identification network to optimize both the verification loss and identification loss. It is supervised by the identification label $t$ and the verification label $s$. On the one hand, a fully-connected layer and the softmax output function are added to map the feature vector of image or video $f$ into a $k$-dim vector $(\hat{p_1}, \hat{p_2}, \cdots, \hat{p_k})$, which represents the predicted possibility of the input belonging to the corresponding identity and $k$ is the total of the identities. Given an input pair, the network simultaneously predicts the IDs of the two input and the similarity score. In the similar fashion to conventional multi-class recognition approaches, we can predict the person using the standard cross-entropy loss as identification loss $L_i$, which is defined as follows:

\begin{equation}
L_i=\sum_{i=1}^k{-p_iln(\hat{p_{i}})}
\end{equation}

where $\hat{p_{i}}$ is the predicted probability of the identity $i$ and the $p_i$ is the identification label, i.e. $p_i=0$ for all i except $p_t=1$, $t$ is the target identity.

As for the verification part, the image and video features $f_i$, $f_v$ are supervised by verification loss. We would like the feature vectors from the same person to be close and the vectors from different people to be widely separated. A non-parameter layer called Square Layer suggested by \citep{10} is added to compare the two feature representations, which takes two feature tensors as inputs and outputs a tensor after substracting and squaring element-wisely. The Square Layer is defined as  

\begin{equation}
f_s=(f_i-f_v)^2 
\end{equation}

where $f_s$ is the output of the Square Layer. After that, a fully-connected layer and the softmax output function are added to map the resulting tensor $f_s$ into a 2-dim vector ($\hat{q_i}$, $\hat{q_2}$), which represents the predicted probability of the input image and video belonging to the same identity, and $q_1+q_2=1$. Pedestrian verification is treated as a binary classification task in this work and adopt the cross-entropy loss as the verification loss $L_v$, which is 

\begin{equation}
L_v=\sum_{i=1}^2{-q_iln(\hat{q_{i}})}
\end{equation}

If the input pair belongs to the same person, $\hat{q_1}=1$, $\hat{q_2}=0$; otherwise, $\hat{q_1}=0$, $\hat{q_2}=1$

What's more, we have found the jointly training for identification and verification loss is crucial for convergence. The overall training loss L for a single pair of input image and video can be defined as follows:

\begin{equation}
L=L_v + \frac{1}{2}L_{ii}+\frac{1}{2}L_{iv}
\end{equation}

where $L_{ii}$ is the identification loss with regard to image and $L_{iv}$ is the identification loss for video.

\section{Experiment}
\label{exp}

In this section, we mainly evaluate the effectiveness of our proposed framework on the image-to-video person re-identification task. The datasets are first introduced, followed by experimental setting and evaluation protocol, and the comparison with several state-of-the-art image-to-video person re-identification is also be presented. 

\subsection{Datasets}
We mainly verify the proposed framework on two different datasets, which are PRID 2011\citep{31} and ILIDS-VID\citep{32}.

\textbf{PRID-2011 dataset} The PRID-2011 dataset consists images sequences recorded from two different, static surveillance cameras, and shows 385 and 749 persons respectively. The first 200 persons are observed in both camera views. Each sequence has variable-length image frames, ranging from 5 to 675, with an average number of 84. This dataset has obvious color changes and shadows in one of the views. Similar to the protocol used in \citep{4}, we only consider the first 200 persons who appear in both cameras.

\textbf{ILDS-VID dataset} The ILDS-VID dataset comprises 600 image sequences of 300 different pedestrians, with one pair of image sequences observed across two disjoint camera views in public open space for each person. Each image sequence has variable length ranging from 23 to 192 image frames, with an average number of 73. It's generally believed that iLIDS-VID dataset is more challenging than PRID 2011 dataset due to clothing similarities among people, lighting and viewpoint variations across camera views, cluttered background and random occlusions. 
 \begin{table*}[!t]
\label{T1}
\caption{Results of our proposed framework and the state-of-the-art methods on the PRID-2011 and ILIDS-VID datasets. The best results are marked with bold symbol}
\centering
\begin{tabular}{c|cccc|cccc}
\hline
Methods & \multicolumn{4}{c|}{PRID-2011(\%)}  & \multicolumn{4}{c}{ILIDS-VID(\%)} \\
\hline
CMC Rank & 1 & 5 & 10 & 20 & 1 & 5 & 10 & 20\\
\hline
Color+DVR & 41.8 & 63.8 & 84.5 & 88.3 & 32.7 & 56.5 & 73.2 & 77.4\\ 
SDALF+DVR & 31.6 & 58 & 81.3 & 85.3 & 26.7 & 49.3 & 65.4 & 71.6\\ 
BoW+XQDA & 31.8 & 58.5 & 78.2 & 81.9 & 14.0 & 32.2 & 52.1 & 59.5\\ 
RCN & 54.3 & 73.4 & 88.5 & 92.5 & 28.8 & 57.7 & 69.6 & 81.9\\ 
RFA-Net & 68.3 & 81.1 & 92.7 & 96.8 & 39.6 & 65.8 & 78.3 & 85.3\\ 
\hline
ours & \textbf{67.5} & \textbf{83.6} & \textbf{93.6} & \textbf{97.6} & \textbf{40.1} & \textbf{67.2} & \textbf{79.7} & \textbf{86.7}\\ 
\hline
\end{tabular}
\end{table*}
\subsection{Experimental setting and evaluation protocol}

Following the setting of \citep{4}, the data is randomly split into 50\% of persons for training and 50\% of the persons for testing for these experiments on the dataset of PRID-2011 and ILDS-VID. All experiments are repeated 10 times with different train/test splits and the results averaged to ensure stable results. As a preprocessing step, images and frames of video are resized to 299*299, satisfying the input requirement of Inception v3 network.

To evaluate the efficiency of the proposed framework, we compare our framework with several state-of-the-art image-to-video person re-identification approaches.Four descriptors including color, Hog3D(orientation Histograms of 3D gradient orientations)\citep{36}, SDALF(Symmetry-Driven Accumulation of Local Features)\citep{33} and BoW(Bag-of-Words)\citep{34}, and two metric learning method including DVR(Discriminative Video fragments selection and Ranking )\citep{31} and XQDA(Cross-view Quadratic Discriminant Analysis)\citep{13} are compared. We also list the results of video-based person re-identification methods RCN(recurrent convolutional network)\citep{4} and RFA-Net(Recurrent Feature Aggregation Network\citep{5}. As the original models of these two methods are based on the matching of video-to-video, we modified their models according to our setting for image-to-video person re-id so that the results reported in this paper are different from those in their original papers.

As we focus on the image-to-video person re-identification task, where the probe is an image and the gallery consists of videos. During the training on PRID-2011 dataset and ILDS-VID dataset, we set the image under the single-shot folder and images folder respectively as the input image, and the video of the other camera view under the multi-shot folder and images folder respectively as the input video. The same strategy is adopted during the test. The network is trained for 500 epochs using stochastic gradient descent with a learning rate of 1e-3, alternating between showing the network positive and negative input pairs. A full epoch consists of showing all positive image-to-video pairs and an equal number of negative pairs, randomly sampled from all training persons. All the experiments are conducted on a NVIDIA GeForce GTX 1080 Ti GPUs server.

In experiments, the matching of each pedestrian's probe image(captured by one camera) with the gallery sequence(captured by other camera) is ranked. When evaluating re-id algorithms, the cumulative matching characteristics(CMC) is usually used. CMC represents the probability that a probe identity appears in different-sized candidate lists. This measurement is acceptable, in practice, when people care more about the returning the ground truth match in the top positions of the rank list. Therefore, to reflect the statistics of the ranks of true matches, the average CMC over 10 trails is adopted as the evaluation metric.Specially, in the testing phase, the similarity between probe image features and those of gallery sequences are computed firstly. Then for each probe image,  a rank list of all the videos in the gallery is sorted in a descending order. Finally, the percentage of true matches found among the first $m$ ranked persons is computed and denoted as CMC rank-m.

\subsection{Experimental results}

With the above experimental setting and evaluation protocol, we obtain the image-to-video person re-identification results using our proposed framework on the PRID-2011 and ILIDS-VID dataset. To ensure a fair comparison, the experiments by compared methods are based on the same datasets and same training/testing splits as ours. As shown in the Table 1, our framework is compared with several state-of-the-art methods from the literature in terms of CMC rank accuracy. 

From the Table 1, we can see that our proposed framework outperforms all compared methods and achieves a CMC rank-1 accuracy of 67.5\% and 40.1\% on PRID-2011 and ILIDS-VID datasets respectively. And RFA-Net has the best result among the compared methods and is comparable to our framework, followed by RCN, which shows the utility of DNNs in the re-identification context, as has been demonstrated in many other application fields. Besides, the experimental results on the ILIDS-VID dataset are generally inferior to those on the PRID-2011 dataset, because PRID-2011 dataset is less challenging than ILDS-VID dataset due to being captured in uncrowded outdoor scenes with rare occlusions and clean background. Overall, the comparison results indicate that our proposed framework is comparable to the state-of-the-art and demonstrate the effectiveness of our framework for the image-to-video person re-identification.

\section{Conclusion}
\label{conclusion}

Image-to-video person re-identification, which attempts to address the problem where the probe is a single image and the gallery comprises videos of different pedestrians from non-overlapping cameras, is different from the conventional person re-identification based on the homogeneous matching. In this paper, we propose a novel end-to-end framework for image-to-video person re-identification, which resorts to LSTMs and CNNs for extracting visual and spatiotemporal information from image and video, combines the strengths of identification model and verification model to improve the discriminative ability of learned features and reuse cross-model embeddings from different but related tasks to narrow down the ``media gap'' between image and video. The experimental results indicate our framework outperforms all compared methods and demonstrate the effectiveness of our framework for dealing with the image-to-video person re-identification task.

As future work, we may incorporate the feature re-construction approaches of heterogeneous data into our framework to help learned features of image and video to be mapped into a closer space. Besides, a more strict verification loss function can be studied to supervise the network training, for the purpose of improving the discriminative ability of learned feature representations.

\section*{Acknowledgments}
This work is supported by the National Social Science Foundation of China (Grant No: 15BGL048), Hubei Province Science and Technology Support Project (Grant No: 2015BAA072), Hubei Provincial Natural Science Foundation of China (Grant No: 2017CFA012), The Fundamental Research Funds for the Central Universities (WUT: 2017II39GX).

\bibliographystyle{model2-names}
\bibliography{reid}

\end{document}